%% file: main.tex
\documentclass[sigconf]{acmart}

\usepackage{booktabs} 

\setcopyright{none}

\acmDOI{xx.xxx/xxx_x}

\acmISBN{xxx}

\acmConference[To appear in 2022 ACM Symposium on Applied Computing (SAC'22)]{}{}{}
\acmYear{2022}
\copyrightyear{2022}

\acmArticle{4}
\acmPrice{15.00}

\begin{document}
\title{A Comparison of Online Hate on Reddit and 4chan: \\A Case Study of the 2020 US Election}

\renewcommand{\shorttitle}{A Comparison of Online Hate on Reddit and 4chan}

\author{Fatima Zahrah}
\orcid{0000-0001-5353-5875}
\affiliation{
  \institution{Department of Computer Science, University of Oxford}
  \streetaddress{P.O. Box 1212}
  \city{Oxford} 
  \country{UK} 
}
\email{fatima.zahrah@cs.ox.ac.uk}

\author{Jason R. C. Nurse}
\affiliation{
  \institution{School of Computing,\\ University of Kent}
  \city{Canterbury}
  \country{UK}}
\email{j.r.c.nurse@kent.ac.uk}

\author{Michael Goldsmith}
\affiliation{
  \institution{Department of Computer Science, University of Oxford}
  \city{Oxford}
  \country{UK}}
\email{michael.goldsmith@cs.ox.ac.uk}


\begin{abstract}
  The rapid integration of the Internet into our daily lives has led to many benefits but also to a number of new, wide-spread threats such as online hate, trolling, bullying, and generally aggressive behaviours. While research has traditionally explored online hate, in particular, on one platform, the reality is that such hate is a phenomenon that often makes use of multiple online networks. In this article, we seek to advance the discussion into online hate by harnessing a comparative approach, where we make use of various Natural Language Processing (NLP) techniques to computationally analyse hateful content from Reddit and 4chan relating to the 2020 US Presidential Elections. Our findings show how content and posting activity can differ depending on the platform being used. Through this, we provide initial comparison into the platform-specific behaviours of online hate, and how different platforms can serve specific purposes. We further provide several avenues for future research utilising a cross-platform approach so as to gain a more comprehensive understanding of the global hate ecosystem.
\end{abstract}

%
%



\maketitle

\input{main2}

\bibliographystyle{ACM-Reference-Format}
\bibliography{sample-bibliography} 

\end{document}

%% file: main2.tex
\section{Introduction}
\label{introduction}
The past few decades have demonstrated how the Internet is playing an ever-increasing role in daily life and has become an integral part of society. In particular, various digital technologies and online platforms for communication have been rapidly adopted into the home and workplace alike. However, this has also introduced several cyber social challenges as digital platforms have provided an effective medium for spreading hateful content, and thus bring new difficulties for agencies responsible for ensuring the boundaries of acceptable and legal behaviour are not crossed \cite{Houseley2017}. The UK government specifically outlined hateful content as one of the primary forms of illegal content online in their Online Harms Paper \cite{HMGovernmentUK2019}.

Nevertheless, the concept of online hate is still considered a complex phenomenon, with its definition evolving across several theoretical paradigms, disciplines and spanning multiple forms of victimisation \cite{schmidt-wiegand-2017-survey}. Due to this complexity, research into online hate is fragmented throughout numerous disciplines. Despite all these extensive approaches and methods proposed to analyse online hate \cite{Zannettou2020,Bevensee2020}, limited research has investigated how hateful behaviours and content compare and relate across different online platforms \cite{shruti2020}. It has only recently been recognised within academic literature that online hate is not simply an issue for a select few platforms, rather networks of hate are often linked across these platforms, forming a global ‘network of networks’ dynamic \cite{Johnson2019}.

\begin{figure*}[ht]
\centering
    \includegraphics[width=0.9\textwidth]{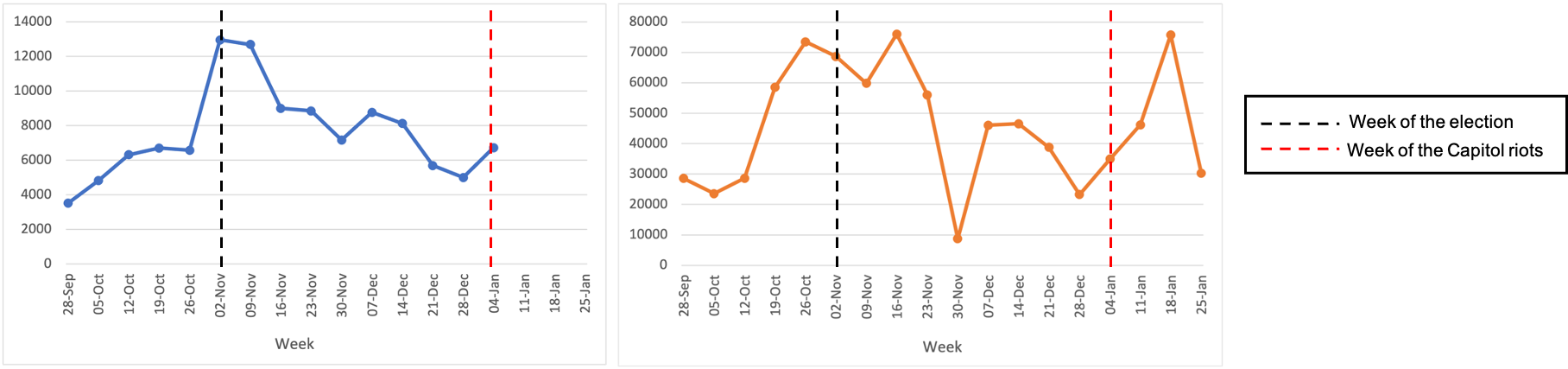}
\caption{Graphs showing the frequency of posts across each dataset: Reddit (left) and 4chan (right).}
\label{fig:participation}
\end{figure*}

Our study applies various computational methods, including topic modelling and sentiment analysis, to explore the type of content that is promoted on Reddit and 4chan to provide unique insight into how various platforms are used within online hate. In particular, this research will make use of data collected over the course of the 2020 US presidential election to investigate how hateful content and narratives compare across both platforms. We use the 2020 election as a case study as previous research has shown that political elections often ``trigger'' hateful discourse online \cite{Scrivens2020}. Through this, we aim to gain an understanding on how online platforms are used for the different functionalities they offer. With our work, this paper:
\begin{itemize}
    \item Examines how the posting behaviour of hateful communities of Reddit and 4chan changes over the course of the 2020 US election and its aftermath.
    \item Identifies the main topics of discussion on both platforms to show how different types of content and narratives are promoted on each platform.
    \item Uses linguistic analysis tools to compare the types of sentiment and levels of emotion used on both platforms.
    \item Analyses the usage of URL domains in posts from both platforms to investigate if any common information sharing takes place and gain a deeper insight into how different platforms are used in distinct ways in online hate.
\end{itemize}

The remainder of the paper will be structured as follows. Section~\ref{methodology} will describe our analysis approach, including the datasets and data-analysis tools that were used. Preliminary results from our findings will be discussed in Section~\ref{results}. We then present our conclusions and outline avenues for future work in Section~\ref{conclusion}.

\section{Methodology}
\label{methodology}
Our comparative analysis of online hate during the 2020 US election on Reddit and 4chan was largely focused on content from white-supremacist ideologies, and was carried out with particular regard to the following research questions:
\begin{itemize}
    \item \textbf{RQ1:} How do the participation and posting trends compare across both platforms over the course of the election and their aftermath?
    \item \textbf{RQ2:} What are the main topics of discussion for hateful users on each of the platforms?
    \item \textbf{RQ3:} Are there any differences in the general sentiment of the posts from both platforms?
    \item \textbf{RQ4:} How do domain-sharing practices compare across both platforms?
\end{itemize}
Our approach therefore comprises three stages: (1) collecting the data from both platforms and observing the posting behaviours, (2) conducting topic modelling on each corpus of posts, and (3) analysing the sentiment of the collected posts to examine their structural properties.

We first carry out an empirical analysis using various computational methods to examine data associated with each respective dataset, including the frequency of posts over time, the most discussed topics as well as keywords, and the different URL domains shared within the posts. The results from this are compared across both datasets to identify any similarities or differences in the composition of their content. This analysis is carried out using the Pandas\footnote{https://pandas.pydata.org/} data analysis library and the `Natural Language Toolkit' (NLTK)\footnote{https://www.nltk.org/} provided by the Python programming language. To identify the main topics of discussion, we conducted topic modelling using the Latent Dirichlet Allocation (LDA) topic detection model with 5 topics, which we found to work better overall than other models, like Non-Negative Matrix Factorization (NMF). We then investigate the sentiment of each set of posts further by using the programmatically coded dictionary from the Linguistic Inquiry and Word Count (LIWC) linguistic analysis tool, which summarises the emotional, cognitive, and structural components in a given text sample \cite{Pennebaker2015}, and thus provides additional avenues for comparison. The results from these are detailed in the subsequent sections.

\subsection{Data Collection}
\label{datacollection}
We chose to analyse Reddit and 4chan within our study as they both represent distinct types of social-media platforms. Reddit is a more mainstream platform that carries out some content moderation, where hateful communities are often cultivated on particular subreddits, but such subreddits have also been removed from the platform if they are increasingly linked to hateful events, such as the subreddits r/fatpeoplehate and r/CoonTown \cite{Chandrasekharan2017}. 4chan, on the other hand, is an anonymous imageboard platform with no content moderation that represents a fringe community with a more specific audience. Both of these online platforms therefore provide a distinct set of functionalities and audiences. We aim to understand the extent to which these platforms can play individual roles and serve different purposes within the wider ecosystem of online hate.

The Reddit dataset was gathered from r/donaldtrump, which has been linked far-right groups and spreading online hate over the past year and was banned earlier this year as a result \cite{Reimann2021}. The 4chan dataset was collected from the Politically Incorrect (/pol/) board, which has also been associated with spreading online hate, and has even been linked to violent acts of extremism including the 2015 Christchurch shooting \cite{Johnson2019}. Both datasets from Reddit and 4chan were collected over the same time period relevant to the timeline of the US elections; we collected data from 1 October 2020 -- 31 January 2021 (so as to include content over the course of the presidential debates, the actual election as well as the presidential certification). We further filtered both datasets using specific terms and keywords in order to collect content specifically related to the election, for instance ``maga'' (Make America Great Again) and ``trump''. The sizes of the two datasets are as follows:
 Reddit dataset: 112,981 posts and 4chan dataset: 1,086,053 posts.

\section{Results and Discussion}
\label{results}
\subsection{Participation Trends}
\label{participation}

With each dataset, we first explore the frequency of content being posted over the course of the collection period during the US election, in answer to RQ1. In Figure~\ref{fig:participation}, we can immediately see that the amount of content posted on 4chan is considerably greater than the content posted on Reddit. Somewhat similar trends in the amount of participation can still be seen, though, across all platforms over the course of the election time frame, where post frequency was measured weekly. Notably, a significant peak in the number of posts can be seen in the weeks during the election (in the week beginning November 2nd), and again in the first week of January following Joe Biden's presidential certification and the resulting Capitol riots \cite{Dave2021}. The r/donaldtrump subreddit was banned as a result of the part it played during these events (hence the abrupt end to the graph points in Figure~\ref{fig:participation}).

\begin{figure}[t]
\centering
    
    \includegraphics[width=0.9\columnwidth]{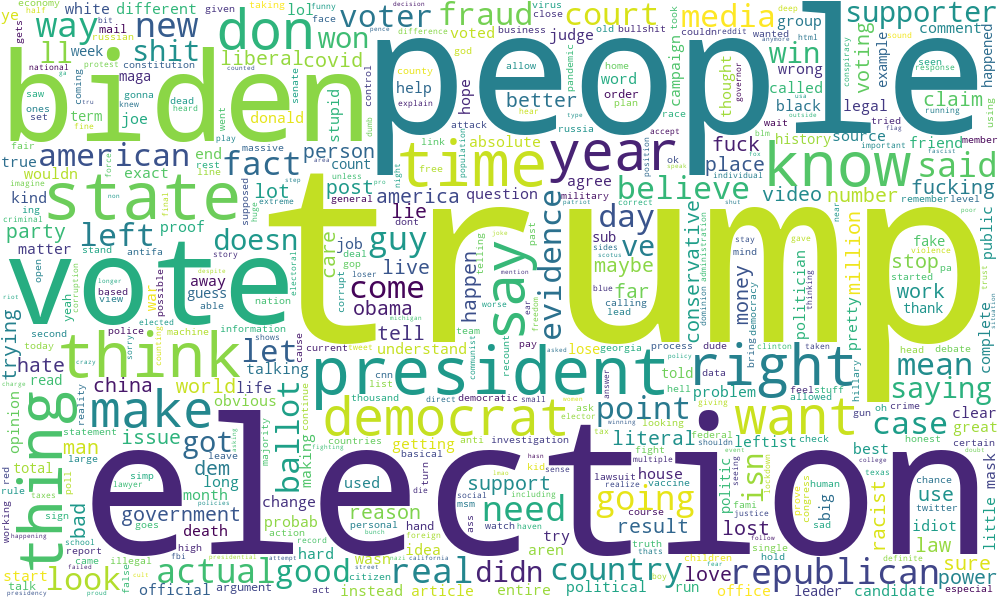}\par\medskip
    \includegraphics[width=0.9\columnwidth]{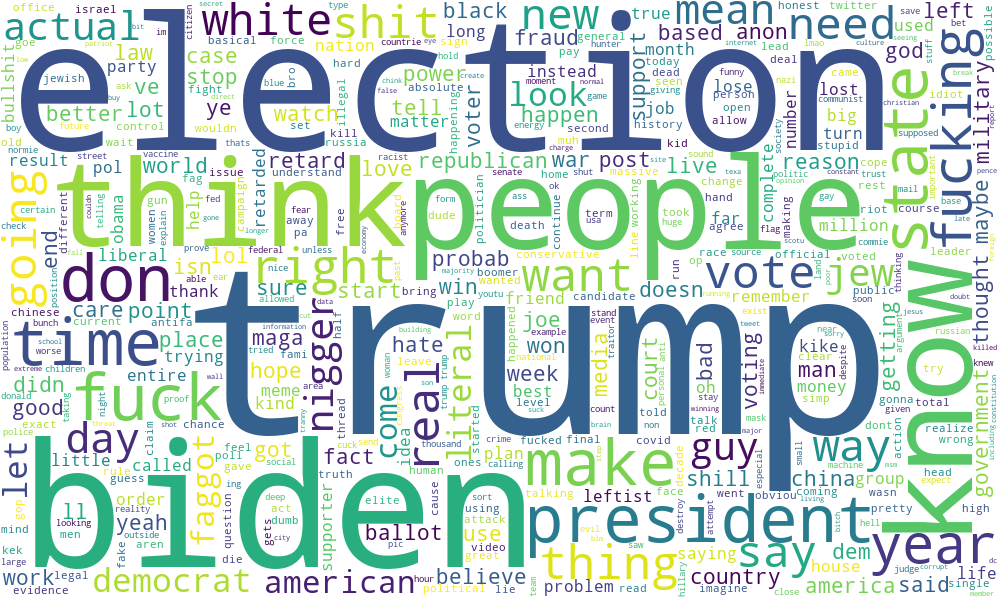}
    
\caption{Word clouds of the most commonly used words across both datasets: Reddit (above) and 4chan (below).}
\label{fig:wordcloud}
\end{figure}

\subsection{Keywords and Topic Analysis}
\label{keyword}
To answer RQ2, we next determine which words and topics were discussed the most in each dataset. In both the Reddit and 4chan datasets, the most common words were, unsurprisingly, \textit{election}, \textit{Trump} and \textit{Biden}. Though, it's worth noting that 4chan posts often refer to groups of ``others'', such as \textit{jew}, and often use offensive and hateful names for them. This suggests hate is more explicit on 4chan than Reddit. Contrastingly, Reddit mentions \textit{fraud}, \textit{media} and \textit{evidence} more frequently.

A topic model also provided further insight into the most discussed subjects within each dataset, with the top five identified topics and the percentage of posts containing them being listed in Table~\ref{table:1}; to get this percentage, the dominant topic out of the five topics was extracted in each post, and a cumulative total of the number of posts for each topic was calculated and represented as a percentage of the total posts in the dataset. When comparing the topics from all the datasets, we can see that similar topics are generally discussed or mentioned. The topic model shows that both sets of posts mention election and voter fraud, though this seems to be discussed more heavily in the Reddit posts. ``MAGA 2020'' and ``MAGA forever'' are also mentioned in both datasets. Most notably though, the Reddit posts also discuss ``election defense'' and make mention of a ``Trump march'', likely referring to the logistical planning of the January capitol riots. Similar to what was shown in the word clouds, the 4chan posts evidently make use of more explicit and derogatory language than any of the other datasets of posts. The topic model shows that such terms seem to especially be used to discuss Trump losing the presidential election (Topic\#2), suggesting such language is used more when voicing frustration.

\begin{table}[t]

\centering
\footnotesize
\caption{A topic model of the most discussed topics and the percentage of posts containing them.}
\begin{tabular}{| c | p{0.19\textwidth} | p{0.19\textwidth} |}
\hline
 & \textbf{Reddit} & \textbf{4chan}\\
\hline
\textit{Topic\#1} & donald trump, support, register, vote, state vote \textbf{(20\%)} & MAGA, awoo, awoo, MAGA hat, MAGA forever, MAGA 2020 \textbf{(25\%)}\\
\hline
\textit{Topic\#2} & make, report voter, trump campaign, fraud, voter fraud \textbf{(15\%)} & still, supporter, vote, lost, lose, going, trump, f***, n****** \textbf{(21\%)}\\
\hline
\textit{Topic\#3} & trump 2020, MAGA, liber tears, MAGA 2020, breaks \textbf{(17\%)} & vote, count, case, state, election fraud, votes, voting, election day \textbf{(23\%)}\\
\hline
\textit{Topic\#4} & election defense, contact state, trump march, stop washington \textbf{(23\%)} & lost, trump lost, lost election, lol, lost biden, white\textbf{(19\%)}\\
\hline
\textit{Topic\#5} & vote, ballot, trump, president, election, voter, people, state \textbf{(25\%)} & watch, president, video, MAGA, youtube, ballot, border, capitol, trump \textbf{(11\%)}\\
\hline
\end{tabular}
\label{table:1}
\end{table}

\subsection{Sentiment Analysis using LIWC}
\label{liwc}
In addition to identifying the main topics of discussion on each platform, we were also interested in examining and comparing the sentiment of the posts. To gain insight into this, we make use of the LIWC linguistic analysis tool to highlight any key differences between each dataset, the findings from which aim to answer RQ3. The results from this analysis are summarised in Table~\ref{table:2}, where the mean percentage of all words within each set of posts that fall into a particular LIWC category is shown. Example words of each category can be found in \cite{Pennebaker2015}. 

We measured the overall sentiments of each platform by using all the LIWC sentiment categories \textit{positive emotion}, \textit{negative emotion}, \textit{anger} and \textit{anxiety}. The results from our LIWC analysis show that both platforms generally used more positive emotion than negative emotion, with posts from Reddit using positive emotion the most. To gain further understanding of the context in which positive emotion was used, we manually examined a sample of posts as well. From this, we were able to find that Reddit posts encouraging users to vote and ``support'' Trump seemed to use more positive emotion. Since we saw in the topic modelling in Section~\ref{keyword} that these were the most prominent topics identified in the Reddit dataset, it is clear how positive emotion is much more present than negative emotion on this platform.

\begin{table}[h]

\centering
\footnotesize
\caption{Results from the linguistic analysis using LIWC.}
\begin{tabular}{| c | c | c |}
\hline
\textbf{LIWC Category} & \textbf{   Reddit   } & \textbf{   4chan   }\\
\hline
\textit{Positive Emotion} & 3.25 & 3.05\\
\hline
\textit{Negative Emotion}& 1.76 & 3.01\\
\hline
\textit{Anxiety} & 0.17 & 0.25\\
\hline
\textit{Anger} & 0.81 & 1.61\\
\hline
\end{tabular}
\label{table:2}
\end{table}

On the other hand, we can also clearly note from the LIWC analysis that negative emotion was used significantly more in the 4chan posts than any other dataset. From assessing a sample of the posts using negative sentiment, we find that this emotion was mainly used when expressing frustration over Trump's loss in the election. The topic modelling we conducted also showed that discussing Trump's loss was a prominent topic of discussion on this platform. Additionally, we can see similar results when looking at the level of \textit{anger} in each dataset from the LIWC results, where 4chan posts made use of such language much more than any of the other platforms.

\subsection{URL Analysis}
To answer our final research question, RQ4, we explored which domains were posted most frequently, where the 10 most popular domains in both datasets have been listed in Table~\ref{table:3}. The difference in the number of URLs used within the Reddit and 4chan posts is quite significant. The 4chan dataset contained 226,394 URLs in total, where 40,256 of these were unique URLs and 21\% of all the posts containing a URL. The posts mostly link to Trump-supporting domains or storage sites such as Pastebin and Flickr. The Reddit dataset also predominantly consists of various links to Trump-supporting domains, including fundraising initiatives as well as voter information and redirection. However, it makes use of a considerably larger amount of URLs. More specifically, 202,445 URLs were used, with 7601 unique URLs, meaning 175\% of these posts used a URL, where posts likely contained multiple URLs, or certain posts would mass-post URLs. This suggests that Reddit users were much more `evidence-based' in their efforts to support Trump.

\begin{table}[h]

\centering
\footnotesize
\caption{The most posted URL domains and their frequency across both datasets.}
\smallskip\begin{tabular}{|p{0.45\columnwidth}|p{0.45\columnwidth}|}
\hline
\textbf{Reddit} & \textbf{4chan}\\
\hline
\textit{www.reddit.com}: 34,786 & \textit{pastebin.com}: 7120\\
\hline
\textit{discord.gg}: 24,973 & \textit{www.promiseskept.com}: 3601\\
\hline
\textit{secure.winred.com}: 19,139 & \textit{www.donaldjtrump.com}: 3590\\
\hline
\textit{www.armyfortrump.com}: 9518 & \textit{www.magapill.com}: 3588\\
\hline
\textit{vote.donaldjtrump.com}: 9515 & \textit{www.whitehouse.gov}: 3584\\
\hline
\textit{vote.gov}: 9513 & \textit{www.flickr.com}: 3376\\
\hline
\textit{www.usa.gov}: 9513 & \textit{www.cbp.gov}: 3371\\
\hline
\textit{trumpvictory.com}: 9513 & \textit{www.armyfortrump.com}: 1861\\
\hline
\textit{share.djt.app}: 9513 & \textit{trumpvictory.com}: 1856\\
\hline
\textit{shop.donaldjtrump.com}: 9513 & \textit{publicpool.kinja.com}: 1678\\
\hline
\end{tabular}
\label{table:3}
\end{table}

\section{Conclusions and Future Work}
\label{conclusion}
Recent events and academic literature have noted that online hate is a phenomenon that incorporates multiple platforms, though little work has been done to understand how these platforms are used strategically. Ultimately, our paper provides some initial comparison across hate-specific online environments on two different platforms (Reddit and 4chan) within the context of the 2020 US presidential elections. Through conducting various computational and linguistic analysis, we observed how the frequency of online participation changes over the time frame of social events relating to real-time offline developments. We also found how content-posting behaviours can change across platforms, particularly in terms of posting URLs from specific domains. The insights we gained from our findings show how different online platforms can serve distinct purposes and play specific roles in the wider ecosystem of online hate. For instance, this paper shows how more mainstream platforms, like Reddit, can be used to encourage their audience to take specific actions, like voting for Trump. We can therefore propose that counter-hate initiatives must be tailored to specific platforms to work effectively. However, further work is needed to build on these observations. 

Namely, extended analysis of multiple other platforms is needed to truly understand and validate whether online content is often specific to the platform is was posted on. It would also be important to gain insight into whether multiple platforms are used in conjunction to achieve some goal or further the reach of hateful content. In this sense, analysing how networks of hate map across several different platforms would provide a more comprehensive and novel perspective of cross-platform online hate. Finally, our findings are only representative of one case study (the US election), therefore future work should compare these across other case studies or events to gain a more comprehensive understanding of how these platforms are used in distinct ways.

%% file: main.bbl

\begin{thebibliography}{12}


\ifx \showCODEN    \undefined \def \showCODEN     #1{\unskip}     \fi
\ifx \showDOI      \undefined \def \showDOI       #1{#1}\fi
\ifx \showISBNx    \undefined \def \showISBNx     #1{\unskip}     \fi
\ifx \showISBNxiii \undefined \def \showISBNxiii  #1{\unskip}     \fi
\ifx \showISSN     \undefined \def \showISSN      #1{\unskip}     \fi
\ifx \showLCCN     \undefined \def \showLCCN      #1{\unskip}     \fi
\ifx \shownote     \undefined \def \shownote      #1{#1}          \fi
\ifx \showarticletitle \undefined \def \showarticletitle #1{#1}   \fi
\ifx \showURL      \undefined \def \showURL       {\relax}        \fi
\providecommand\bibfield[2]{#2}
\providecommand\bibinfo[2]{#2}
\providecommand\natexlab[1]{#1}
\providecommand\showeprint[2][]{arXiv:#2}

\bibitem[\protect\citeauthoryear{Bevensee, Aliapoulios, Dougherty, Baumgartner,
  Mccoy, and Blackburn}{Bevensee et~al\mbox{.}}{2020}]%
        {Bevensee2020}
\bibfield{author}{\bibinfo{person}{Emmi Bevensee}, \bibinfo{person}{Maxwell
  Aliapoulios}, \bibinfo{person}{Quinn Dougherty}, \bibinfo{person}{Jason
  Baumgartner}, \bibinfo{person}{Damon Mccoy}, {and} \bibinfo{person}{Jeremy
  Blackburn}.} \bibinfo{year}{2020}\natexlab{}.
\newblock \showarticletitle{{SMAT : The Social Media Analysis Toolkit}}.
\newblock \bibinfo{journal}{\emph{Proceedings of the International AAAI
  Conference on Web and Social Media Workshops}}  \bibinfo{volume}{14}
  (\bibinfo{year}{2020}).
\newblock


\bibitem[\protect\citeauthoryear{Chandrasekharan, Pavalanathan, Srinivasan,
  Glynn, Eisenstein, and Gilbert}{Chandrasekharan et~al\mbox{.}}{2017}]%
        {Chandrasekharan2017}
\bibfield{author}{\bibinfo{person}{Eshwar Chandrasekharan},
  \bibinfo{person}{Umashanthi Pavalanathan}, \bibinfo{person}{Anirudh
  Srinivasan}, \bibinfo{person}{Adam Glynn}, \bibinfo{person}{Jacob
  Eisenstein}, {and} \bibinfo{person}{Eric Gilbert}.}
  \bibinfo{year}{2017}\natexlab{}.
\newblock \showarticletitle{You Can't Stay Here: The Efficacy of Reddit's 2015
  Ban Examined Through Hate Speech}.
\newblock \bibinfo{journal}{\emph{Proc. ACM Hum.-Comput. Interact.}}
  \bibinfo{volume}{1} (\bibinfo{year}{2017}), 22.
\newblock
\urldef\tempurl%
\url{https://doi.org/10.1145/3134666}
\showDOI{\tempurl}


\bibitem[\protect\citeauthoryear{Dave, McNichols, and Sabia}{Dave
  et~al\mbox{.}}{2021}]%
        {Dave2021}
\bibfield{author}{\bibinfo{person}{Dhaval~M Dave}, \bibinfo{person}{Drew
  McNichols}, {and} \bibinfo{person}{Joseph~J Sabia}.}
  \bibinfo{year}{2021}\natexlab{}.
\newblock \bibinfo{booktitle}{\emph{Political Violence, Risk Aversion, and
  Non-Localized Disease Spread: Evidence from the U.S. Capitol Riot}}.
\newblock \bibinfo{type}{Working Paper} 28410. \bibinfo{institution}{National
  Bureau of Economic Research}.
\newblock
\urldef\tempurl%
\url{https://doi.org/10.3386/w28410}
\showDOI{\tempurl}


\bibitem[\protect\citeauthoryear{{Department for Digital Culture Media and
  Sport (UK)}}{{Department for Digital Culture Media and Sport (UK)}}{2019}]%
        {HMGovernmentUK2019}
\bibfield{author}{\bibinfo{person}{{Department for Digital Culture Media and
  Sport (UK)}}.} \bibinfo{year}{2019}\natexlab{}.
\newblock \bibinfo{title}{{Online Harms White Paper}}.
\newblock
\newblock
\urldef\tempurl%
\url{https://www.gov.uk/government/consultations/online-harms-white-paper}
\showURL{%
\tempurl}


\bibitem[\protect\citeauthoryear{Housley, Webb, Edwards, Procter, and
  Jirotka}{Housley et~al\mbox{.}}{2017}]%
        {Houseley2017}
\bibfield{author}{\bibinfo{person}{William Housley}, \bibinfo{person}{Helena
  Webb}, \bibinfo{person}{Adam Edwards}, \bibinfo{person}{Rob Procter}, {and}
  \bibinfo{person}{Marina Jirotka}.} \bibinfo{year}{2017}\natexlab{}.
\newblock \showarticletitle{Membership categorisation and antagonistic Twitter
  formulations}.
\newblock \bibinfo{journal}{\emph{Discourse \& Communication}}
  \bibinfo{volume}{11}, \bibinfo{number}{6} (\bibinfo{year}{2017}),
  \bibinfo{pages}{567--590}.
\newblock
\urldef\tempurl%
\url{https://doi.org/10.1177/1750481317726932}
\showDOI{\tempurl}
\showeprint{https://doi.org/10.1177/1750481317726932}


\bibitem[\protect\citeauthoryear{Johnson, Leahy, Restrepo, Velasquez, Zheng,
  Manrique, Devkota, and Wuchty}{Johnson et~al\mbox{.}}{2019}]%
        {Johnson2019}
\bibfield{author}{\bibinfo{person}{N.~F. Johnson}, \bibinfo{person}{R Leahy},
  \bibinfo{person}{N~Johnson Restrepo}, \bibinfo{person}{N Velasquez},
  \bibinfo{person}{M Zheng}, \bibinfo{person}{P Manrique}, \bibinfo{person}{P
  Devkota}, {and} \bibinfo{person}{S Wuchty}.} \bibinfo{year}{2019}\natexlab{}.
\newblock \showarticletitle{{Hidden Resilience and Adaptive Dynamics of the
  Global Online Hate Ecology}}.
\newblock \bibinfo{journal}{\emph{Nature}} \bibinfo{volume}{573},
  \bibinfo{number}{7773} (\bibinfo{year}{2019}), \bibinfo{pages}{261--265}.
\newblock
\showISSN{1476-4687}
\urldef\tempurl%
\url{https://doi.org/10.1038/s41586-019-1494-7}
\showDOI{\tempurl}


\bibitem[\protect\citeauthoryear{Pennebaker, Boyd, Jordan, and
  Blackburn}{Pennebaker et~al\mbox{.}}{2015}]%
        {Pennebaker2015}
\bibfield{author}{\bibinfo{person}{James Pennebaker}, \bibinfo{person}{Ryan
  Boyd}, \bibinfo{person}{Kayla Jordan}, {and} \bibinfo{person}{Kate
  Blackburn}.} \bibinfo{year}{2015}\natexlab{}.
\newblock \bibinfo{booktitle}{\emph{{The Development and Psychometric
  Properties of LIWC2015}}}.
\newblock \bibinfo{type}{{T}echnical {R}eport}.
  \bibinfo{institution}{University of Texas, Austin}.
\newblock


\bibitem[\protect\citeauthoryear{Phadke and Mitra}{Phadke and Mitra}{2020}]%
        {shruti2020}
\bibfield{author}{\bibinfo{person}{Shruti Phadke} {and}
  \bibinfo{person}{Tanushree Mitra}.} \bibinfo{year}{2020}\natexlab{}.
\newblock \showarticletitle{Many Faced Hate: A Cross Platform Study of Content
  Framing and Information Sharing by Online Hate Groups}. In
  \bibinfo{booktitle}{\emph{Proceedings of the 2020 CHI Conference on Human
  Factors in Computing Systems}} \emph{(\bibinfo{series}{CHI ’20})}.
  \bibinfo{publisher}{Association for Computing Machinery},
  \bibinfo{address}{New York, NY, USA}, \bibinfo{pages}{1–13}.
\newblock
\showISBNx{9781450367080}
\urldef\tempurl%
\url{https://doi.org/10.1145/3313831.3376456}
\showDOI{\tempurl}


\bibitem[\protect\citeauthoryear{Reimann}{Reimann}{2021}]%
        {Reimann2021}
\bibfield{author}{\bibinfo{person}{Nicholas Reimann}.}
  \bibinfo{year}{2021}\natexlab{}.
\newblock \bibinfo{title}{{Reddit Bans ‘r/donaldtrump' Subreddit}}.
\newblock
\newblock
\urldef\tempurl%
\url{https://www.forbes.com/sites/nicholasreimann/2021/01/08/reddit-bans-rdonaldtrump-subreddit/?sh=5980347038b3}
\showURL{%
\tempurl}


\bibitem[\protect\citeauthoryear{Schmidt and Wiegand}{Schmidt and
  Wiegand}{2017}]%
        {schmidt-wiegand-2017-survey}
\bibfield{author}{\bibinfo{person}{Anna Schmidt} {and} \bibinfo{person}{Michael
  Wiegand}.} \bibinfo{year}{2017}\natexlab{}.
\newblock \showarticletitle{A Survey on Hate Speech Detection using Natural
  Language Processing}. In \bibinfo{booktitle}{\emph{Proceedings of the Fifth
  International Workshop on Natural Language Processing for Social Media}}.
  \bibinfo{publisher}{Association for Computational Linguistics},
  \bibinfo{address}{Valencia, Spain}, \bibinfo{pages}{1--10}.
\newblock
\urldef\tempurl%
\url{https://doi.org/10.18653/v1/W17-1101}
\showDOI{\tempurl}


\bibitem[\protect\citeauthoryear{Scrivens, Burruss, Holt, Chermak, Freilich,
  and Frank}{Scrivens et~al\mbox{.}}{2020}]%
        {Scrivens2020}
\bibfield{author}{\bibinfo{person}{Ryan Scrivens}, \bibinfo{person}{George~W.
  Burruss}, \bibinfo{person}{Thomas~J. Holt}, \bibinfo{person}{Steven~M.
  Chermak}, \bibinfo{person}{Joshua~D. Freilich}, {and}
  \bibinfo{person}{Richard Frank}.} \bibinfo{year}{2020}\natexlab{}.
\newblock \showarticletitle{Triggered by Defeat or Victory? Assessing the
  Impact of Presidential Election Results on Extreme Right-Wing Mobilization
  Online}.
\newblock \bibinfo{journal}{\emph{Deviant Behavior}} \bibinfo{volume}{0},
  \bibinfo{number}{0} (\bibinfo{year}{2020}), \bibinfo{pages}{1--16}.
\newblock
\urldef\tempurl%
\url{https://doi.org/10.1080/01639625.2020.1807298}
\showDOI{\tempurl}
\showeprint{https://doi.org/10.1080/01639625.2020.1807298}


\bibitem[\protect\citeauthoryear{Zannettou, Finkelstein, Bradlyn, and
  Blackburn}{Zannettou et~al\mbox{.}}{2020}]%
        {Zannettou2020}
\bibfield{author}{\bibinfo{person}{Savvas Zannettou}, \bibinfo{person}{Joel
  Finkelstein}, \bibinfo{person}{Barry Bradlyn}, {and} \bibinfo{person}{Jeremy
  Blackburn}.} \bibinfo{year}{2020}\natexlab{}.
\newblock \showarticletitle{A Quantitative Approach to Understanding Online
  Antisemitism}.
\newblock \bibinfo{journal}{\emph{Proceedings of the International AAAI
  Conference on Web and Social Media}} \bibinfo{volume}{14},
  \bibinfo{number}{1} (\bibinfo{date}{May} \bibinfo{year}{2020}),
  \bibinfo{pages}{786--797}.
\newblock
\urldef\tempurl%
\url{https://ojs.aaai.org/index.php/ICWSM/article/view/7343}
\showURL{%
\tempurl}


\end{thebibliography}
